\documentclass[a4paper,fleqn]{cas-sc}

\usepackage[numbers,sort&compress]{natbib}
\usepackage{graphicx}
\usepackage{amsmath}
\usepackage{amssymb}
\usepackage{amsfonts}
\usepackage{algorithm}
\usepackage{algorithmic}
\usepackage{booktabs}
\usepackage{multirow}
\usepackage{makecell}
\usepackage{array}
\usepackage{threeparttable}
\usepackage{bm}
\usepackage{xspace}
\usepackage[table]{xcolor}
\usepackage{microtype}
\usepackage{listings}
\newcommand{\argmax}{\mathop{\mathrm{arg\,max}}}
\newcommand{\argmin}{\mathop{\mathrm{arg\,min}}}
\newcommand{\R}{\mathbb{R}}
\newcommand{\method}{IAE\xspace}
\definecolor{vermillion}{HTML}{D55E00}
\definecolor{linkblue}{HTML}{0072B2}
\newcommand{\marg}[1]{\textcolor{vermillion}{\textbf{#1}}}
\definecolor{barblue}{HTML}{9CC3E0}
\definecolor{bargrey}{HTML}{C9CCD1}
\newlength{\barunit}
\newcommand{\vbar}[2]{\textcolor{#1}{\rule[0.05ex]{#2\barunit}{0.85ex}}}
\definecolor{bartrack}{HTML}{EFEFEF}
\newcommand{\nbar}[4]{\makebox[1.7em][r]{#1}\,\makebox[5.6em][l]{\scriptsize[#4]}\makebox[\barunit][l]{\rlap{\textcolor{bartrack}{\rule[0.05ex]{\barunit}{0.85ex}}}\vbar{#3}{#2}}}
\newcommand{\sig}{$^{\ast}$}
\newlength{\tabw}
\begin{document}
\let\WriteBookmarks\relax

\shorttitle{Instance-Anchored Interaction Evidence}
\shortauthors{X. Xiao et al.}

\title [mode = title]{Instance-anchored interaction evidence: Grounding robot plans in human pointing and handling}

\author[1]{Xinliang Xiao}[orcid=0009-0001-9670-0575]
\ead{xinliangxiao@njust.edu.cn}

\author[1]{Bowen Yang}[orcid=0009-0008-0229-0060]
\ead{bowenyang@njust.edu.cn}

\author[1]{Wenjing Zhang}[orcid=0009-0000-4262-0820]
\ead{924110800235@njust.edu.cn}

\author[1]{Li Yang}[orcid=0000-0002-8152-7642]
\ead{12025071@njust.edu.cn}

\author[1]{Wei Zhou}[orcid=0000-0003-3225-0576]
\cormark[1]
\ead{weizhou@njust.edu.cn}

\affiliation[1]{organization={Nanjing University of Science and Technology},
            city={Nanjing},
            postcode={210094},
            country={China}}

\cortext[1]{Corresponding author}

\begin{abstract}
A robot that assists people must often act on what a person has shown rather than said: which of several identical cartons was pointed at, or which box was handled. The plan is executed from the final scene, whereas the evidence occurs earlier, possibly on objects that have since moved. We propose instance-anchored interaction evidence (IAE), which registers every object of the final scene to its public identifier, keeps each identity through the video by backward mask propagation, and describes every frame by the geometry between hands, forearms and these instances. An evidence network trained only from task outcomes scores the instances. For pointing tasks, a grammar-constrained dynamic program trained with a structured loss decodes object--destination programs; symbolic programs handle reference disambiguation and, without learning, episodic tasks. On 1{,}255 WatchAct benchmark requests, scored by symbolic execution, IAE reaches 64.2\% plan success on implicit-intent tasks against 27.5\% for a 32B vision-language model (strict success 49.7\% against 15.4\%), and 46.4\% against 27.0\% on restoration, reversal and imitation without task-specific training. Controls with the same perception overlays, the same 32 frames, forward tracking, or a relation model trained on the same labels do not explain the gain. Given IAE's evidence as text with its meaning explained, the same language model reaches 57.4\%: most of the gain comes from the instance-anchored evidence, and the explicit programs add 6.8 points at a fraction of the cost. Pointing remains the hardest case, with 16.9\% strict success. The code is available at \href{https://github.com/WeiZhou96/iae-watchact}{\textcolor{linkblue}{\nolinkurl{https://github.com/WeiZhou96/iae-watchact}}}.
\end{abstract}

\begin{highlights}
\item Final-scene object identities are registered and tracked back through the video
\item Hand--instance evidence is learned from task outcomes without frame labels
\item A grammar-constrained program decoder turns pointing evidence into plans
\item Plan success on WatchAct implicit-intent tasks rises from 27.5\% to 64.2\%
\item The same evidence lifts a 32B VLM planner from 27.5\% to 57.4\% plan success
\end{highlights}


\maketitle

\section{Introduction}
\label{sec:intro}

People frequently instruct others by acting instead of speaking. They point at the carton that should go into the tray, pick up one of two butter boxes and put it back, or rearrange a few objects and expect them to be restored later. For a robot that joins such a task, the video of the person carries information that the language instruction leaves out. The WatchAct benchmark~\cite{watchact} formalizes this setting: each instance pairs a real human-action video and an instruction with a simulated scene, and the robot plan is scored after execution. Its evaluation of current vision-language models (VLMs) shows that the video-to-plan step remains the main bottleneck.

A common approach is to give the video frames and the instruction to a VLM and let it produce the plan directly~\cite{wake2023gpt4v,seedo}. This works when the plan follows from what the scene looks like, but it breaks down when the plan depends on \emph{which} instance a person interacted with. Scenes in these tasks contain several visually identical objects, the plan must refer to them by identifiers defined in the final scene, and the relevant gesture happens seconds earlier, sometimes on an object that has since been moved. Recent studies report that multimodal models ground pointing gestures poorly and are drawn to salient or nearby objects~\cite{egopointbench}, and that they perform close to chance when identical objects must be told apart by tracking them through time~\cite{shellgame}. In our experiments, a 32B VLM reaches 27.5\% plan success on the implicit-intent tasks of WatchAct and solves only 16.5\% of restoration tasks, which require remembering where each object started.

The difficulty lies in the link between an observed interaction and a specific physical instance at execution time. Gesture-aware robot policies add hand skeletons or fingertip rays to the robot's current observation~\cite{give,gesvla}, and egocentric grounding methods resolve a single pointing act to a box~\cite{egopointground,egopointvqa}. In a third-person demonstration, however, the person may issue several deictic cues whose order defines which object goes where, and the identities that the plan needs are only defined at the end of the video. Mapping a whole video to a plan with a VLM leaves both the temporal composition and the identity bookkeeping implicit.

We propose to make this link explicit. Instance-anchored interaction evidence (\method) registers every object of the final scene to its public identifier by fitting the public layout to the detections, propagates instance masks backwards through the video so that each identity is kept while objects are handled and moved, and describes each frame by the geometric relation between the detected hands, forearms and these instances. On top of this representation, simple programs produce the plan: which instances were indicated and in which order for pointing tasks, which instances were selected for reference disambiguation, and which instances moved and where they came from for episodic tasks.

Pointing is the case in which the raw geometry is least reliable. A downward point at a tray from behind the table can land on the basket in the image, and a retracting hand sweeps across other objects. Instead of annotating gesture episodes, we train a small evidence network from task outcomes only. The objects and destinations that appear in the goal of a training activity serve as video-level labels for a multiple-instance loss, and a dynamic program that enforces the deictic grammar of object and destination cues turns the frame-level evidence into a complete program. A structured loss on whole programs, computed with the same decoder, trains the evidence to make the decoded program correct rather than only to rank candidates.

Our contributions are:
\begin{itemize}
  \item An instance-anchored representation of human interaction in demonstration videos that combines public-layout registration, identity-preserving backward tracking and hand--forearm geometry, and that serves pointing, selection and manipulation tasks alike.
  \item A weakly supervised learning scheme for deictic evidence that uses only task outcomes, combining a multiple-instance loss, grammar-constrained program decoding and a program-level structured loss.
  \item An evaluation on 1{,}255 WatchAct requests with official and strict success metrics, activity-grouped and recording-order-blocked cross-validation, forward-tracking, same-supervision and matched-sampling controls, and baselines from two VLM families, including 8B and 32B models that receive the same perception overlays, the evidence of \method itself, or are asked to verify ambiguous decisions.
\end{itemize}
\method reaches 64.2\% plan success on the implicit-intent tasks, 36.7 percentage points above the 32B VLM, and 49.7\% strict success against 15.4\%; on restoration, reversal and imitation it reaches 46.4\% without task-specific training, where the 32B VLM reaches 27.0\%. All plans are evaluated by symbolic execution; we do not evaluate grasping or physical execution.

\section{Related work}
\label{sec:related}

\textbf{Robot planning from human videos.} Several systems derive robot plans or skills from human demonstration videos. GPT-4V has been prompted to convert task videos into executable robot programs~\cite{wake2023gpt4v}, SeeDo combines keyframe selection, perception and VLM reasoning for long-horizon pick-and-place plans~\cite{seedo}, and VLMimic grounds object-centric movements for fine-grained skills~\cite{vlmimic}. Video-conditioned policies such as Vid2Robot~\cite{vid2robot}, XSkill~\cite{xskill} and Track2Act~\cite{track2act} learn to act from demonstrations or predicted point tracks. Benchmarks now evaluate these abilities separately from low-level control: WatchAct~\cite{watchact} measures video-to-plan reasoning over 14 tasks, EgoPlan-Bench~\cite{egoplan} evaluates egocentric planning, and RoboReel~\cite{roboreel} evaluates learning from observation. These systems mostly rely on the VLM to resolve which object was involved. We instead compute this binding explicitly and keep the VLM only as a baseline and a verifier.

\textbf{Understanding pointing and deictic gestures.} DeePoint recognizes pointing and estimates its direction from body joints~\cite{deepoint}. YouRefIt~\cite{yourefit} shows that gestures matter as much as language for embodied reference, and the Touch-Line Transformer~\cite{touchline} constrains the referent to lie on the eye--fingertip line. GIRAF combines gesture detectors with language-model reasoning for table-top assistance~\cite{giraf}, and a gesture pseudo-language combines several types of gestures into sentences that state an action and its parameters, such as the object and its destination, parsed in real time against the object poses that the robot perceives~\cite{vanc2023gesture}. Recent work adds hand skeletons and fingertip rays to vision-language-action policies~\cite{give,gesvla}, feeds hand-keypoint tokens to multimodal models~\cite{egopointvqa}, or benchmarks egocentric pointing resolution~\cite{egopointground,egopointbench,vistaref}. Most of these methods resolve one pointing act in the current view. Gesture sentences do compose several cues, but they are addressed to the robot as commands and interpreted in the scene it currently observes. Our setting differs in three respects that are evaluated together: the cues are observed in a third-person video, objects may move after they were indicated, so the operations must be bound to the instances of the scene at the end of the video, and the evidence model is learned from task outcomes without frame-level labels.

\textbf{Weak supervision and structured decoding over time.} Multiple-instance learning from video-level labels is standard for weakly supervised temporal action localization~\cite{pmil}. Transcript-supervised segmentation trains frame scores through differentiable alignment, for example with discriminative dynamic time warping~\cite{d3tw} or outlier-robust alignment~\cite{dropdtw}. Our decoder is a dynamic program over deictic events, and our supervision is weaker than a transcript: only the final object--destination relations are known, while the order and the timing of the cues are latent.

\textbf{Tracking, identity and program-based reasoning.} Keeping track of objects that are handled, occluded or contained is a known weakness of video models~\cite{shellgame,perceptiontest}. Fine-tuning a VLM to generate object trajectories as explicit intermediate reasoning largely removes this weakness on a shell-game benchmark~\cite{shellgame}; such reasoning keeps the identities for the model's own answer, whereas we bind them to the instances of the execution scene and do not fine-tune a VLM. Dedicated methods maintain object identity across long egocentric videos~\cite{outofsight}. Promptable video segmentation~\cite{sam2} makes such tracking practical for arbitrary objects. Program-based visual reasoning composes perception modules with generated code~\cite{visprog,vipergpt}, and visual prompting overlays marks on images to improve grounding~\cite{som}. \method follows the modular view but fixes the program structure by task and learns only the evidence that the programs consume. Uncertainty-aware planners ask for help when a plan is ambiguous~\cite{knowno,refguard}; we use the decoding margin of our programs to report selective prediction.

\section{Problem formulation}
\label{sec:problem}

Each request consists of a video $V=\{I_t\}_{t=1}^{T}$ of a person acting on a table, an instruction $\ell$, and a public scene description $S=(\mathcal{O},s_0)$. The set $\mathcal{O}$ lists the object identifiers (for example \texttt{milk\_1}, \texttt{milk\_2}, \texttt{basket\_1}), and $s_0$ assigns each object to a region of the table: a $3\times3$ grid of main regions, a fixture strip, container interiors and cabinet drawers. The scene $s_0$ describes the state from which the robot executes. For most tasks this is the final frame of the video; for imitation it is the first frame. The system outputs a set of operations $A=\{(o_j,d_j)\}$ that move object $o_j\in\mathcal{O}$ into region $d_j$. The operations are compiled into pick-and-place actions, executed symbolically, and the resulting state is compared with the goal state.

Two metrics are reported. The official plan success counts a request as solved when every object whose goal differs from $s_0$ ends in its goal region; moving other objects is not penalized. We therefore also report strict success, which additionally requires every other object to stay where it was. Region names in $s_0$ and in the goal are defined in a canonical camera frame, and requests recorded from side or oblique cameras, or phrased relative to the person, are scored after mapping to this frame.

The tasks differ in the evidence they require. In \emph{nonverbal cue} (NC) tasks the person points at objects and destinations, and $A$ must reproduce these object--destination pairs. In \emph{reference disambiguation} (RD) tasks the person selects one or more of several similar objects by pointing, touching or moving them, and the instruction asks the robot to place ``the selected'' or ``the other'' objects into a named container, possibly defined relative to the person. In the episodic tasks \emph{restore}, \emph{reversal} and \emph{imitation} the person moves objects, and the robot must return them to their initial regions or repeat the moves.

\section{Method}
\label{sec:method}

\begin{figure}[pos=!tbp]
  \centering
  \includegraphics[width=\textwidth]{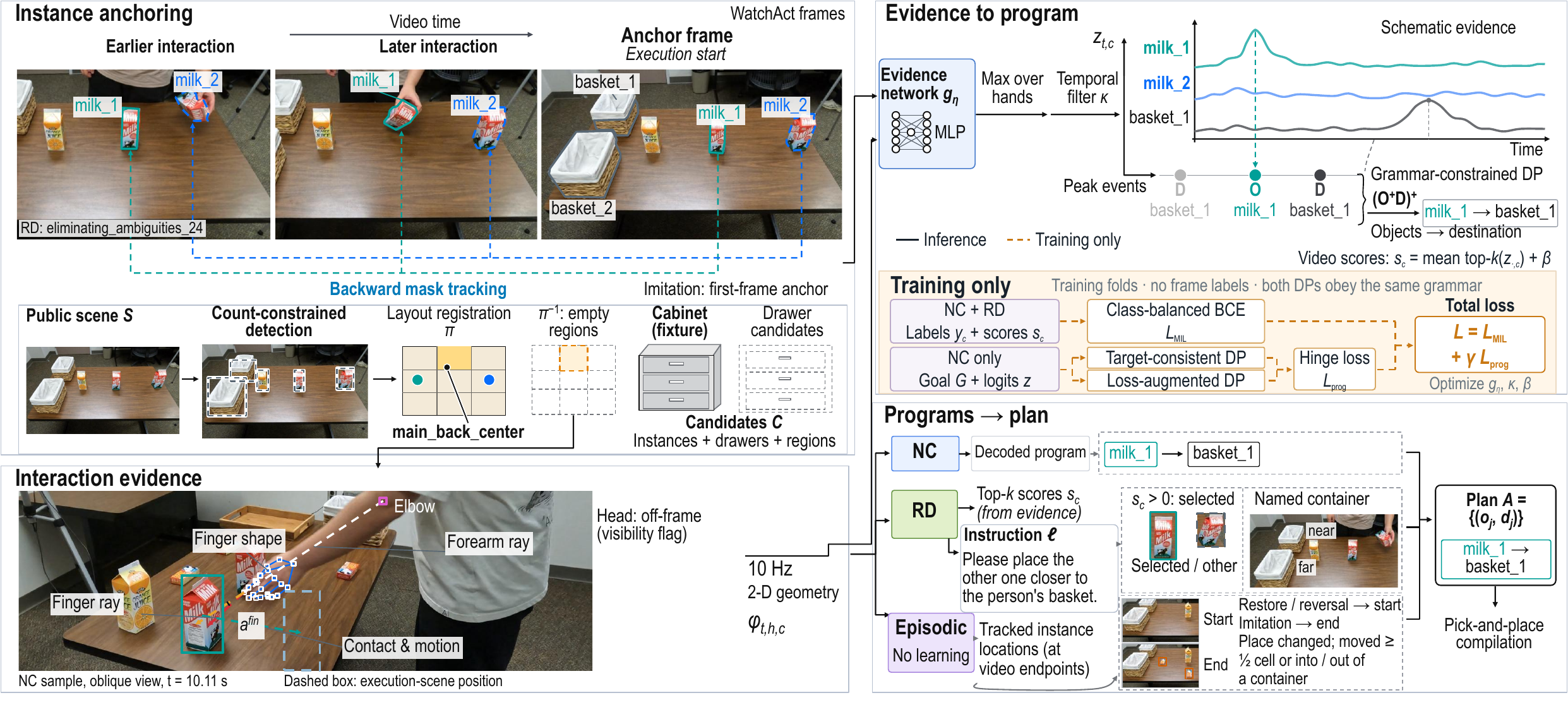}
  \caption{Overview of instance-anchored interaction evidence (\method). Objects of the execution scene are registered to their public identifiers and tracked backwards through the video. Hand and forearm geometry relative to each tracked instance forms the interaction evidence. An evidence network trained from task outcomes scores every candidate in every frame. For pointing tasks, a grammar-constrained dynamic program decodes object--destination operations; reference disambiguation applies a symbolic program to the learned video-level scores, and episodic tasks compare the places of tracked instances at both ends of the video without learning. Images are frames of WatchAct videos (a reference-disambiguation video, a pointing video and, for the episodic branch, the first and last frames of a restoration video); the evidence curves are schematic. Solid arrows denote inference and dashed orange arrows training-only paths.}
  \label{fig:framework}
\end{figure}

Fig.~\ref{fig:framework} summarizes \method. Section~\ref{sec:anchor} registers and tracks every instance, Section~\ref{sec:evidence} builds hand--instance evidence, Section~\ref{sec:learning} learns deictic evidence from task outcomes, and Section~\ref{sec:programs} describes the programs that map evidence to operations.

\subsection{Final-scene registration and identity-preserving tracking}
\label{sec:anchor}

\textbf{Count-constrained detection.} The public scene fixes how many instances of each category exist. We query an open-vocabulary detector~\cite{groundingdino} once per category on the anchor frame (the last frame, or the first frame for imitation), pool the candidate boxes, and greedily select exactly the public number of boxes per category in decreasing score order, suppressing any box that overlaps an already selected box with an intersection over union of at least $0.5$. This removes frequent confusions between similar packages such as butter and cream-cheese boxes.

\textbf{Layout registration.} Detections must be mapped to identifiers such as \texttt{milk\_2}. Each public object that stands on the table has a canonical coordinate $q_o\in\R^2$ given by the column and row of its grid region, with the fixture strip at $x=-1$. For a detection with image box $b$ we use its ground point $p(b)$, the bottom centre of the box. The image-to-table map is
\begin{equation}
  \pi(p)=R_\theta\,\mathrm{diag}(\alpha_x,\alpha_y)\,p+\tau ,
  \label{eq:pi}
\end{equation}
with rotation $R_\theta$, anisotropic scale $(\alpha_x,\alpha_y)$ and translation $\tau$. Let $\Sigma$ be the set of bijections $\sigma$ that assign the detections of each category to the public identifiers of that category. Registration solves
\begin{equation}
  (\hat\sigma,\hat\pi)=\argmin_{\sigma\in\Sigma,\;\theta\in\Theta_v,\;\alpha,\tau}\;\sum_{o\in\mathcal{O}_{\mathrm{tab}}}\big\|\pi\big(p(b_{\sigma(o)})\big)-q_o\big\|^2 ,
  \label{eq:register}
\end{equation}
where $\mathcal{O}_{\mathrm{tab}}$ contains the objects standing on the table. For fixed $\sigma$ and $\theta$ the problem is linear in $(\alpha,\tau)$ and is solved by least squares; $\theta$ is searched on a grid $\Theta_v$ whose range is restricted by the public camera perspective $v$ for front and side cameras, and $\Sigma$ is enumerated exhaustively (at most $2\times10^4$ assignments). The inverse $\hat\pi^{-1}$ projects the centre $q_r$ of every table region $r$ into the image, so that empty regions can serve as candidate destinations. Registration uses only the public layout, which all baselines receive as well.

\textbf{Backward tracking.} From the registered boxes on the anchor frame, instance masks are propagated through the whole video with a promptable video segmenter~\cite{sam2}, moving away from the anchor frame. Each track keeps the identifier of its execution-scene instance while the object is carried, occluded by the hand or partly hidden in a container. A cabinet is split into three drawer candidates by dividing its box vertically.

\subsection{Hand--instance interaction evidence}
\label{sec:evidence}

Frames are sampled at 10\,Hz. A detector localizes the person, 21 hand keypoints are estimated inside the person box~\cite{mediapipehands}, and a body-pose model provides the elbows and head keypoints~\cite{rtmpose}. The candidate set $\mathcal{C}$ contains the tracked instances, the drawer candidates and the projected table regions. Let $x_{t,c}$ be the image centre of candidate $c$ in frame $t$, $B_{t,c}$ its box, $f_{t,h}$ the index fingertip of hand $h$, $u_{t,h}$ the unit direction from the index knuckle to the fingertip, $\varepsilon_{t,h}$ the elbow of the same arm, $\zeta_t$ the mean head keypoint, $W$ the image width, and $\angle(a,b)=\arccos\big(a^{\top}b/(\|a\|\|b\|)\big)$. With the offset $\Delta_{t,h,c}=x_{t,c}-f_{t,h}$ and the forearm direction $v_{t,h}=(f_{t,h}-\varepsilon_{t,h})/\|f_{t,h}-\varepsilon_{t,h}\|$, the central geometric quantities are, omitting the indices $t,h,c$,
\begin{equation}
\begin{aligned}
  a^{\mathrm{fin}}&=\angle(u,\Delta), & \rho&=\|\Delta\|/W,\\
  a^{\mathrm{arm}}&=\angle(v,\Delta), & a^{\mathrm{head}}&=\angle(f-\zeta,\Delta),\\
  \delta^{\perp}&=|v\times\Delta|/W, & \delta^{\parallel}&=v^{\top}\Delta/W,\\
  m&=\|x_{t,c}-x_{t_a,c}\|/W, & \chi&=\mathbf{1}\big[\exists j\!\in\!\mathcal{J}:k_{j}\in B^{+}\big],
\end{aligned}
\label{eq:geom}
\end{equation}
where $\times$ is the 2-D cross product, $\mathcal{J}$ indexes the thumb, index and middle fingertips $k_{t,h,j}$, $B^{+}_{t,c}$ is the box enlarged by 15\% of its size, and $t_a$ is the anchor frame. The feature vector $\phi_{t,h,c}\in\R^{28}$ groups them as follows:
\begin{itemize}
  \item finger shape: straightness of the index finger (cosine between its proximal and distal segments), its extension relative to the palm, flexion of the other fingers, and the hand-detection confidence;
  \item finger ray: $a^{\mathrm{fin}}$, $\rho$ and $\log\rho$, the distance relative to the size of $B_{t,c}$, the vertical offset of $x_{t,c}$ from the fingertip, the ranks of both among all candidates, and whether $c$ is the nearest candidate within the cone $a^{\mathrm{fin}}<25^\circ$;
  \item forearm and head rays: $a^{\mathrm{arm}}$, $\delta^{\perp}$, $\delta^{\parallel}$ and $a^{\mathrm{head}}$, each with a flag that indicates whether the elbow or the head is visible;
  \item contact and motion: $\chi$, the displacement $m_{t,c}$ of $c$ from its execution-scene position, its speed, and the fingertip speed;
  \item context: the candidate type (movable object, container, drawer or region) and the normalized time $t/T$, together with a constant input.
\end{itemize}
The forearm and head rays follow the observation that people align the target with the eye--fingertip line~\cite{touchline}. The image of a 3-D line through the fingertip is a 2-D line through the fingertip, so these rays constrain the image position of the target even though its depth is unknown. The two hand slots are ordered by handedness so that a slot refers to the same hand over time.

\subsection{Learning deictic evidence from task outcomes}
\label{sec:learning}

\textbf{Evidence network.} A two-layer perceptron $g_\eta$ with 48 hidden units scores each hand--candidate pair from $\phi_{t,h,c}$ and a one-hot task indicator $\boldsymbol{\iota}_{\mathrm{task}}$. The maximum over the visible hands gives a frame logit, and a learnable temporal filter $\kappa$ of seven frames, initialized as a moving average, smooths it:
\begin{equation}
  z_{t,c}=\Big(\kappa * \max_{h\in\mathcal{H}_{\cdot}} g_\eta\big(\phi_{\cdot,h,c},\boldsymbol{\iota}_{\mathrm{task}}\big)\Big)_t ,
  \label{eq:z}
\end{equation}
where $\mathcal{H}_t$ is the set of hands detected in frame $t$ and frames without a hand receive a constant low logit. The video-level score of candidate $c$ pools its $k$ largest frame logits,
\begin{equation}
  s_c=\frac{1}{k}\sum_{t\in\mathrm{top}_k(z_{\cdot,c})}z_{t,c}+\beta ,
  \label{eq:sc}
\end{equation}
with a learned bias $\beta$ and $k=5$ frames, which need not be contiguous.

\textbf{Multiple-instance labels.} For an NC training activity, the objects and destinations that appear in its goal were indicated at some unknown time, and all other candidates were never indicated; this assumes that demonstrations contain no retracted or superseded cues. This yields candidate labels $y_c\in\{0,1\}$ without frame annotation. For RD, the selected set is recovered from the goal through the instruction: ``the selected'' objects are the goal objects, and ``the other'' objects are the remaining instances of the reference domain, which is the category of the goal objects or, when that category has no further instance, all movable objects. For RD instructions that refer to ``that container'', the goal container is labelled as indicated and the other containers as not indicated. In RD, the remaining candidates (regions, drawers, and containers unless the instruction refers to ``that container'') are not supervised and receive weight $w_c=0$; all other candidates have $w_c=1$. With $\mathrm{sigmoid}(\cdot)$ the logistic function and $w^{+}=N^{-}/N^{+}$ the ratio of negative to positive labels among the supervised candidates ($w_c=1$) of the training folds, the multiple-instance loss is the class-balanced binary cross-entropy
\begin{equation}
\begin{aligned}
  \mathcal{L}_{\text{MIL}}=-\sum_{c\in\mathcal{C}}w_c\Big[&w^{+}y_c\log\mathrm{sigmoid}(s_c)\\
  &+(1-y_c)\log\big(1-\mathrm{sigmoid}(s_c)\big)\Big].
\end{aligned}
  \label{eq:mil}
\end{equation}

\textbf{Grammar-constrained decoding.} A pointing demonstration is a sequence of object cues followed by a destination cue, possibly repeated. The deictic grammar is therefore $(O^{+}D)^{+}$, and all objects preceding a destination cue are assigned to it. Candidate events $e=(t,c,r)$ are the local maxima of $z_{\cdot,c}$,
\begin{equation}
  \mathcal{E}_c=\Big\{t:\ z_{t,c}=\max_{|t'-t|\le 2}z_{t',c},\ \ z_{t,c}\ge\max_{t'}z_{t',c}-3\Big\},
  \label{eq:events}
\end{equation}
of which at most three per candidate, at least one second apart, are kept in decreasing order of $z_{t,c}$. The role is $r=O$ for movable objects and $r=D$ for containers, drawers and regions. For the time-sorted events $e_1,\dots,e_n$, a program $\omega$ is a subsequence that is accepted by the grammar, uses distinct objects and at most $N_{\max}=3$ objects, and has score
\begin{equation}
  F(\omega)=\sum_{e\in\omega}\big(z_{e}-\lambda\big).
  \label{eq:decode}
\end{equation}
The maximizer is found exactly by the recursion
\begin{equation}
\begin{aligned}
  Q(i,U,P)=\max\Big\{&Q(i{+}1,U,P),\\
    &z_{e_i}-\lambda+Q\big(i{+}1,U,P\cup\{c_i\}\big)\ \ [\mathrm{obj}],\\
    &z_{e_i}-\lambda+Q\big(i{+}1,U\cup P,\emptyset\big)\ \ [\mathrm{dst}]\Big\},
\end{aligned}
  \label{eq:dp}
\end{equation}
where $U$ is the set of objects already assigned to a destination and $P$ the set of objects awaiting one. The option [obj] is admissible only if $r_i=O$, $c_i\notin U\cup P$ and $|U\cup P|<N_{\max}$, and [dst] only if $r_i=D$ and $P\neq\emptyset$; inadmissible options take the value $-\infty$. The recursion terminates with $Q(n{+}1,U,P)=0$ if $P=\emptyset$ and $U\neq\emptyset$, and $-\infty$ otherwise. Since $|U\cup P|\le 3$, the number of states is polynomial in the number of events. The recursion is exact over the candidate events; times outside the retained peaks are not considered. Because the terminal condition excludes the empty program, a non-empty plan is returned whenever a valid program exists, and abstention is left to selective prediction (Section~\ref{sec:results}). Algorithm~\ref{alg:decode} summarizes the procedure. The decoded program determines at once which candidates were indicated and how objects pair with destinations. We use $\lambda=0$ throughout.

\begin{algorithm}[!tbp]
  \caption{Grammar-constrained program decoding}
  \label{alg:decode}
  \begin{algorithmic}[1]
    \REQUIRE frame logits $z\in\R^{T\times|\mathcal{C}|}$, candidate roles, cost $\lambda$, $N_{\max}$
    \ENSURE operations $A$, program score $F^{\ast}$, margin $\mu$
    \STATE $\mathcal{E}\leftarrow\emptyset$
    \FORALL{candidates $c$ with role $r_c\in\{O,D\}$}
      \STATE add the top three peaks of $\mathcal{E}_c$ (Eq.~\eqref{eq:events}), at least 10 frames apart, as events $(t,c,r_c)$ to $\mathcal{E}$
    \ENDFOR
    \STATE sort $\mathcal{E}$ by time into $e_1,\dots,e_n$
    \STATE evaluate $Q(1,\emptyset,\emptyset)$ by Eq.~\eqref{eq:dp} with memoization over $(i,U,P)$, storing the maximizing choice
    \IF{$Q(1,\emptyset,\emptyset)=-\infty$}
      \RETURN $A=\emptyset$, $F^{\ast}=-\infty$, $\mu=0$ (no valid program)
    \ENDIF
    \STATE backtrack the stored choices to the selected events $\omega^{\ast}$; $F^{\ast}\leftarrow Q(1,\emptyset,\emptyset)$
    \STATE $A\leftarrow\emptyset$; $P\leftarrow\emptyset$
    \FORALL{$e=(t,c,r)\in\omega^{\ast}$ in temporal order}
      \IF{$r=O$}
        \STATE $P\leftarrow P\cup\{c\}$
      \ELSE
        \STATE $A\leftarrow A\cup\{(o,c):o\in P\}$; $P\leftarrow\emptyset$
      \ENDIF
    \ENDFOR
    \STATE $F^{-}\leftarrow$ best score by Eq.~\eqref{eq:dp} over states $(i,U,P,b)$, where $b$ marks a pair outside $A$, with terminal states that satisfy $b$ or $U\neq\{o:(o,d)\in A\}$
    \STATE $\mu\leftarrow F^{\ast}-F^{-}$ (or $F^{\ast}$ if $F^{-}=-\infty$)
    \RETURN $A$, $F^{\ast}$, $\mu$
  \end{algorithmic}
\end{algorithm}

\textbf{Program-level structured loss.} The multiple-instance loss only asks each candidate to be indicated at some time, whereas the decoder depends on when and in which order the scores peak. We therefore add a structured hinge loss computed with the same dynamic program. Let $G$ be the goal operations of a training video and $A(\omega)$ the operations of program $\omega$. The task loss counts wrong and missing operations,
\begin{equation}
  \Delta(A,G)=|A\setminus G|+|G\setminus A|,
  \label{eq:taskloss}
\end{equation}
so that an object sent to a wrong destination costs two units, an extra object one and a missed object one. Let $\omega^{\star}$ be the best program with $A(\omega^{\star})=G$. It is decoded by Eq.~\eqref{eq:dp} with three restrictions: object events are limited to the goal objects, a destination event is admissible only if it is the goal destination of every pending object, and the terminal condition requires that all goal objects have been assigned. When the candidate events admit no such program, the video contributes only $\mathcal{L}_{\text{MIL}}$; this applies to 15--16\% of the NC training videos in each run. On the held-out logits of the ensemble, the 20 of 115 NC videos without such a program split evenly between a goal object or destination that is not a candidate, because it was not registered, and the restriction to three peaks per candidate; in none of them does the order of the peaks rule the program out. The loss-augmented program is
\begin{equation}
  \hat\omega=\argmax_{\omega}\Big(F(\omega)+\Delta\big(A(\omega),G\big)\Big).
  \label{eq:augment}
\end{equation}
Since each destination event assigns all pending objects, and the pending set is part of the state, $\Delta-|G|$ decomposes over destination events: closing the pending set $P$ with destination $d$ adds $\sum_{o\in P}\big(\mathbf{1}[(o,d)\notin G]-\mathbf{1}[(o,d)\in G]\big)$. Eq.~\eqref{eq:augment} is therefore solved exactly by the same recursion. Then
\begin{equation}
  \mathcal{L}_{\text{prog}}=\max\Big(0,\;F(\hat\omega)+\Delta\big(A(\hat\omega),G\big)-F(\omega^{\star})\Big),
  \label{eq:prog}
\end{equation}
which is the margin-rescaled structured hinge with the same task loss in the search and in the penalty, and gradients reach the frame logits selected by both programs. On 3{,}000 random event sequences of up to twelve events, the recursion returned the optimum of exhaustive enumeration for free, goal-constrained and loss-augmented decoding and for the margin of Eq.~\eqref{eq:margin}. The full objective is
\begin{equation}
  \mathcal{L}=\mathcal{L}_{\text{MIL}}+\gamma\,\mathcal{L}_{\text{prog}} ,
  \label{eq:total}
\end{equation}
where $\mathcal{L}_{\text{prog}}$ is summed over the NC training videos that admit a goal-consistent program. In the implementation, Eq.~\eqref{eq:mil} is averaged over the candidates of a video, both losses are summed over the training videos, and the total is divided by their number. Training minimizes $\mathcal{L}_{\text{MIL}}$ for 150 epochs and then $\mathcal{L}$ with $\gamma=0.1$ for 20 epochs.

\subsection{Programs over instance-anchored evidence}
\label{sec:programs}

\textbf{Nonverbal cues.} The operations $A$ of the decoded program $\omega^{\ast}$ are the output. Its margin
\begin{equation}
  \mu=F(\omega^{\ast})-\max_{\omega:\,A(\omega)\neq A(\omega^{\ast})}F(\omega)
  \label{eq:margin}
\end{equation}
serves as the confidence for selective prediction. If no program with different operations exists, the empty program, with score 0, serves as the alternative, so that $\mu=F(\omega^{\ast})$; a video without any valid program receives no operations and $\mu=0$. The margin is a ranking score, not a calibrated probability, and it is negative when the only valid program has a negative score. The maximum is computed exactly with the recursion of Eq.~\eqref{eq:dp}, whose state is extended by a flag $b$ that records whether a pair outside $A(\omega^{\ast})$ has been produced. Because objects are distinct, a program reproduces $A(\omega^{\ast})$ exactly when, at the end, $b$ is unset and $U$ equals the objects of $A(\omega^{\ast})$; the terminal condition admits all other programs (Algorithm~\ref{alg:decode}).

\textbf{Reference disambiguation.} Let $\mathcal{M}\subset\mathcal{C}$ be the movable instances and $c^{+}=\argmax_{c\in\mathcal{M}}s_c$. The selected set and the reference domain are
\begin{equation}
\begin{aligned}
  \hat S&=\{c\in\mathcal{M}:s_c>0\},\\
  \mathcal{D}&=\{c\in\mathcal{M}:\mathrm{cat}(c)=\mathrm{cat}(c^{+})\},
\end{aligned}
  \label{eq:rdsel}
\end{equation}
where $\hat S=\{c^{+}\}$ if no score is positive, and $\mathcal{D}$ backs off to $\mathcal{M}$ when it contains a single instance. The instruction determines whether the selected objects or ``the other'' objects are meant, and whether one or several; the target set is
\begin{equation}
  \mathcal{T}=\begin{cases}
    \{c^{+}\} & \text{selected, singular},\\
    \hat S & \text{selected, plural},\\
    \{\argmin_{c\in\mathcal{D}}s_c\} & \text{other, singular},\\
    \mathcal{D}\setminus\hat S & \text{other, plural}.
  \end{cases}
  \label{eq:rdtarget}
\end{equation}
The destination is the container named in the instruction. When several containers of that category exist and the instruction is relative to the person, the person's median ground point $\bar p$ over the video is mapped to the table and
\begin{equation}
  \hat d=\argmin_{k\in\mathcal{K}}\ \pm\big\|q_k-\hat\pi(\bar p)\big\|
  \label{eq:rddest}
\end{equation}
selects the closer ($+$) or farther ($-$) container among the matching containers $\mathcal{K}$; for ``that container'', the container with the highest $s_c$ is chosen. The output is $A=\{(o,\hat d):o\in\mathcal{T}\}$, and the confidence is $\min_{c\in\mathcal{M}}|s_c|$, a heuristic that measures how far the scores lie from the selection threshold but not how ambiguous the ranking of the candidates or the destination is.

\textbf{Episodic tasks.} Restoration, reversal and imitation reuse the registration and the tracks without learning. Let $t_a$ be the anchor end of the video and $t_b$ the opposite end (the first frame for restoration and reversal, the last for imitation). The place $\psi_o(t)$ of instance $o$ is a container when its centre lies inside a tracked container box, a drawer when it lies inside a cabinet box, and otherwise the table region that contains $\hat\pi(p(b_{t,o}))$; it is read at the nearest frame within 1.5\,s of each end in which $o$ is visible, and an instance that stays invisible is assigned to the container nearest to its first or last visible position. With the displacement in grid cells
\begin{equation}
  \Delta_o=\big\|\hat\pi\big(p(b_{t_a,o})\big)-\hat\pi\big(p(b_{t_b,o})\big)\big\|,
  \label{eq:disp}
\end{equation}
where $b_{t_a,o}$ and $b_{t_b,o}$ are the boxes at the frames at which the places are read, the program moves every instance whose place changed,
\begin{equation}
  A=\big\{\big(o,\psi_o(t_b)\big):\ \psi_o(t_b)\neq\psi_o(t_a)\ \wedge\ \big(\Delta_o\ge\tfrac12\ \vee\ \nu_o\big)\big\},
  \label{eq:episodic}
\end{equation}
where $\nu_o$ indicates that $\psi_o(t_a)$ or $\psi_o(t_b)$ is a container or a drawer, so that moves into and out of containers count regardless of their image displacement. For restoration and reversal this returns each moved instance to its initial place, and for imitation it moves each instance to its final place.

\section{Experimental setup}
\label{sec:setup}

\textbf{Data.} We use five WatchAct tasks~\cite{watchact}. The implicit-intent tasks are nonverbal cue (39 activities, 195 requests) and reference disambiguation (52 activities, 260 requests). The episodic tasks are restore (57 activities, 285 requests), reversal (45, 225) and imitation (58, 290). Each activity is recorded by a front, a side and an oblique camera, and each request combines one video with a camera-centred or person-centred spatial reference. One NC video is corrupted in the released data; its request is kept in the denominator and counted as a failure, as are all requests for which a method returns no valid plan. For one further NC video, from the side camera, registration finds too few detections, so \method returns no plan for its two requests, which also count as failures; the learned NC experiments therefore use 115 of the 117 NC videos.

\textbf{Protocol.} The evidence network is evaluated by five-fold cross-validation in which all videos and requests of an activity fall into the same fold. Labels are derived from the goals of training activities only, and test activities are processed without access to goals. We also report a recording-order-blocked split in which activities are ordered by the file number of their front video and cut into five contiguous blocks. File order serves as a proxy for recording sessions, which are not annotated; recordings made close in time tend to share a fold, but participants and rooms are not controlled. Reported numbers are means over three training seeds, and the ensemble averages the frame logits of the three seeds. The episodic programs involve no learning and are applied to all requests directly.

\textbf{Metrics.} We report the official plan success rate (SR) of the benchmark, and strict success, which also requires all non-target objects to remain in place. Confidence intervals are 95\% intervals from 5{,}000 bootstrap resamples of activities, and paired comparisons use a two-sided sign test on per-activity differences. We call a difference significant when its bootstrap interval excludes zero and report the sign test, which ignores the size of the per-activity differences, alongside.

\begin{table}[pos=!tbp]
  \centering\scriptsize\setlength{\barunit}{26pt}
  \renewcommand{\arraystretch}{1.1}
  \caption{Implicit-intent tasks: plan success rate (SR) and strict success (\%) on all 455 requests.}
  \label{tab:main}
  \begin{threeparttable}
    \begin{tabular*}{\textwidth}{@{\extracolsep{\fill}}l cc cc l l c@{}}
      \toprule
      & \multicolumn{2}{c}{Nonverbal cue (195)} & \multicolumn{2}{c}{Reference disamb. (260)} & \multicolumn{2}{c}{All (455)} & \\
      \cmidrule(lr){2-3}\cmidrule(lr){4-5}\cmidrule(lr){6-7}
      Method & SR & Strict & SR & Strict & \multicolumn{1}{c}{SR [95\% CI]\tnote{a}} & \multicolumn{1}{c}{Strict [95\% CI]} & $\Delta$SR\tnote{b} \\
      \midrule
      Qwen3-VL-8B, direct & 16.9 & 4.1 & 27.7 & 20.8 & \nbar{23.1}{0.231}{bargrey}{16.3, 30.5} & \nbar{13.6}{0.136}{bargrey}{7.9, 20.0} & $-4.4$ \\
      Qwen3-VL-32B, direct & 22.6 & 4.6 & 31.2 & 23.5 & \nbar{27.5}{0.275}{bargrey}{20.9, 34.5} & \nbar{15.4}{0.154}{bargrey}{9.9, 21.1} & 0.0 \\
      Qwen3-VL-8B + overlays & 25.6 & 7.2 & 34.6 & 27.3 & \nbar{30.8}{0.308}{bargrey}{23.1, 38.9} & \nbar{18.7}{0.187}{bargrey}{12.3, 25.3} & $+3.3$ \\
      Qwen3-VL-32B + overlays & 21.5 & 3.1 & 34.2 & 24.2 & \nbar{28.8}{0.288}{bargrey}{22.0, 36.0} & \nbar{15.2}{0.152}{bargrey}{9.7, 21.1} & $+1.3$ \\
      InternVL3.5-8B, direct & 10.8 & 4.1 & 34.6 & 17.3 & \nbar{24.4}{0.244}{bargrey}{18.5, 31.0} & \nbar{11.6}{0.116}{bargrey}{7.3, 16.5} & $-3.1$ \\
      Qwen3-VL-32B + evidence\tnote{d} & 37.4 & $\mathbf{17.4}$ & 25.4 & 20.8 & \nbar{30.5}{0.305}{bargrey}{23.1, 38.5} & \nbar{19.3}{0.193}{bargrey}{13.0, 26.2} & $+3.0$ \\
      \quad + frames & 29.7 & 12.8 & 26.2 & 18.8 & \nbar{27.7}{0.277}{bargrey}{20.9, 34.7} & \nbar{16.3}{0.163}{bargrey}{11.0, 22.0} & $+0.2$ \\
      \quad explained\tnote{e} & 35.9 & 12.3 & 73.5 & 68.5 & \nbar{57.4}{0.574}{bargrey}{49.0, 65.3} & \nbar{44.4}{0.444}{bargrey}{35.8, 52.7} & $+29.9$ \\
      Hand-crafted evidence & 2.6 & 2.6 & 75.4 & 70.0 & \nbar{44.2}{0.442}{bargrey}{34.9, 53.6} & \nbar{41.1}{0.411}{bargrey}{31.4, 50.5} & $+16.7$ \\
      \midrule
      \method, single model\tnote{c} & \underline{37.6} & 15.7 & \underline{78.7} & \underline{73.2} & \nbar{\underline{61.1}}{0.611}{barblue}{52.5, 69.2} & \nbar{\underline{48.6}}{0.486}{barblue}{39.8, 57.7} & $+33.6$ \\
      \method, three-seed ensemble & $\mathbf{42.6}$ & \underline{16.9} & $\mathbf{80.4}$ & $\mathbf{74.2}$ & \nbar{$\mathbf{64.2}$}{0.642}{barblue}{55.4, 72.5} & \nbar{$\mathbf{49.7}$}{0.497}{barblue}{40.4, 58.9} & \marg{$+36.7$} \\
      \bottomrule
    \end{tabular*}
    \begin{tablenotes}
      \item[a] \scriptsize Bars start at zero and reach full length at 100\% (blue: \method; grey: baselines). Intervals: 5{,}000 activity-cluster bootstrap resamples; for the single model they are computed on seed-averaged outcomes.
      \item[b] \scriptsize Difference in overall SR to Qwen3-VL-32B direct. For the ensemble, the 95\% interval of the paired difference is $[+27.0,+45.9]$ and $p=10^{-7}$ (sign test over activities; Table~\ref{tab:paired}).
      \item[c] \scriptsize Mean over three seeds; standard deviations of overall SR and strict success across seeds are 1.5 and 1.0.
      \item[d] \scriptsize The VLM receives the held-out evidence of the \method ensemble as text (NC: candidate events; RD: evidence scores and person--container distances) and produces the plan; ``+ frames'' also attaches the 32 frames.
      \item[e] \scriptsize Evidence as text, with a prompt that explains what the scores mean and that ``the other one'' refers to the instances not selected (Appendix~\ref{app:prompts}).
    \end{tablenotes}
  \end{threeparttable}
\end{table}

\textbf{Baselines.} Direct planning gives Qwen3-VL-8B or Qwen3-VL-32B~\cite{qwen3vl} 32 uniformly sampled frames (longest side 896 pixels) together with the instruction and the public scene expressed in the spatial frame of the request, including the region descriptions of that frame, and asks for object--region pairs; predictions are mapped back to the canonical frame before scoring. The overlay variants draw the same perception outputs used by \method on every frame: tracked boxes labelled with public identifiers, hand skeletons and forearm lines. The hand-crafted variant replaces the evidence network by a fixed score that combines pointing shape, ray angle and distance for NC and uses the maximum displacement of each instance for RD, and applies the same programs. The 32B model is run with FP8 weights on a single GPU. To test whether the gap is specific to one model family, we also run InternVL3.5-8B~\cite{internvl35} with the same frames, prompt and output format; following its standard video setting, each frame is encoded as a single 448-pixel tile. Finally, to separate the evidence of \method from its programs, the 32B model receives the held-out evidence of the \method ensemble as structured text and produces the plan itself, once with the plain prompt and once with a prompt that explains the evidence (Appendix~\ref{app:prompts}): for NC the candidate events of Eq.~\eqref{eq:events} with time, public identifier, role and logit, and for RD the video-level score of every instance and container and the distance of the person to each container. The prompt, scene description and output format are otherwise those of direct planning, with or without the 32 frames. Both variants use the FP8 model of direct planning.

\textbf{Implementation.} \method uses Grounding~DINO~\cite{groundingdino} for detection, SAM~2.1 with the tiny backbone~\cite{sam2} for tracking, MediaPipe Hands~\cite{mediapipehands} and RTMPose~\cite{rtmpose} for keypoints. The evidence network has 48 hidden units, a temporal filter of 7 frames and 28 input features. It is trained with Adam (learning rate $3\times10^{-3}$, weight decay $10^{-4}$) on full batches. All thresholds of the programs are fixed a priori: decoding cost $\lambda=0$, selection threshold $s_c>0$, and a minimum displacement of half a grid cell for episodic moves.

\section{Results}
\label{sec:results}

\begin{figure}[pos=!tbp]
  \centering
  \includegraphics[width=\textwidth]{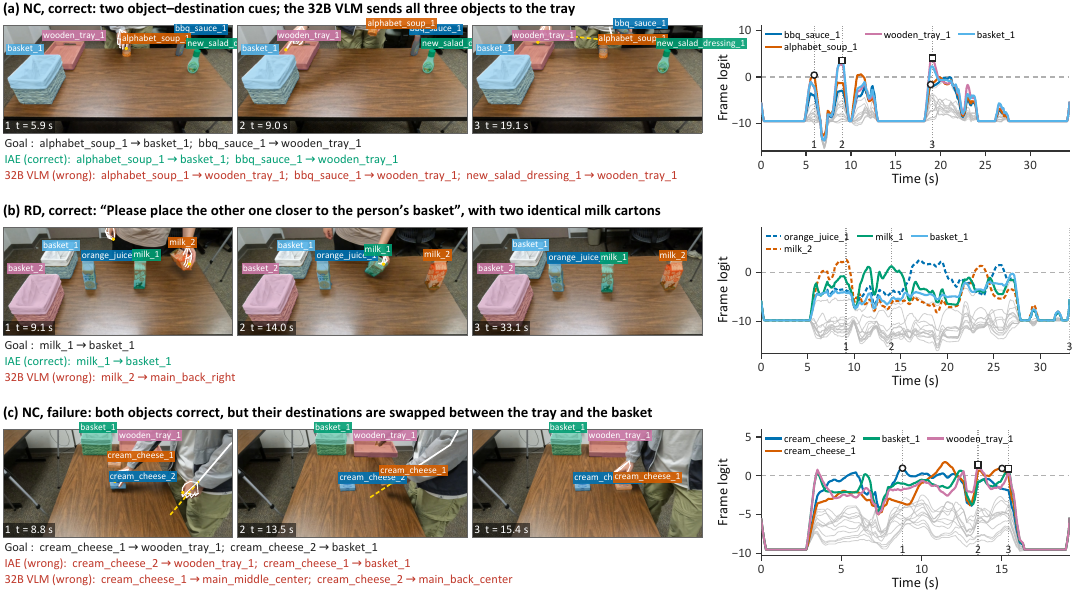}
  \caption{Examples from the test folds. Frames show the instance masks and public identifiers obtained by registration and backward tracking, the hand landmarks (white, index fingertip in yellow) and the forearm line (yellow, dashed) that \method computes at that frame; frames are cropped at the same height for all examples. Right: frame logits of the three-seed ensemble; coloured solid curves are candidates in the goal, coloured dashed curves are candidates decoded but not in the goal (NC) or the other movable instances (RD), grey curves are the remaining candidates, and circles and squares mark decoded object and destination events (NC only). Numbers link frames to time. NC frames are the first object event and the first and last destination events; RD frames are the evidence peaks of the two cartons and the final frame. Examples were drawn at random (fixed seed) from three predefined groups: NC requests from the front camera solved strictly by \method and not by the 32B VLM (12 requests), NC requests with the correct objects and a tray--basket confusion (18), and RD requests solved strictly by \method and not by the 32B VLM in which the selected instance moves by more than 60 pixels (6).}
  \label{fig:cases}
\end{figure}

\subsection{Implicit-intent tasks}

Table~\ref{tab:main} compares \method with direct planning. All three VLMs reach below 32\% SR overall; scaling Qwen3-VL from 8B to 32B adds 4.4 points, which is not significant ($p=0.07$), and InternVL3.5-8B, from a different model family, reaches 24.4\%, so the gap is not specific to one family. Their strict success on NC stays below 5\%: when the models commit to a plan, they usually move additional objects. \method reaches 61.1\% SR with a single model and 64.2\% with the ensemble. The improvement over the 32B model holds on both tasks and for every seed; against the strongest VLM baseline, the 8B model with overlays, the single models gain between 28.6 and 32.3 points ($p<10^{-5}$).

The overlay baselines show that access to the same perception is not sufficient. Drawing identifiers, hand skeletons and forearm lines helps the 8B model by 7.7 points but not the 32B model, and neither approaches \method. The hand-crafted variant isolates the other side of the comparison. With the same registration, tracking and programs but no learned evidence, RD already reaches 75.4\%, which shows that selection is largely solved once each instance keeps its identity. For NC the hand-crafted evidence fails almost completely (2.6\%), and the learned evidence raises SR to 37.6\%.

\textbf{Can a VLM use the evidence of \method?} The evidence rows of Table~\ref{tab:main} give the 32B model the evidence that the programs of \method consume, and the model produces the plan. With the plain prompt, it reaches 37.4\% SR and 17.4\% strict success on NC, close to the ensemble decoded by the grammar (42.6\% and 16.9\%; paired difference in SR $5.1$ points, $[-5.6,15.9]$), but only 25.4\% on RD: in 191 of its 194 RD failures it selects the wrong objects, typically the most strongly handled instance when the instruction asks for the other one. A prompt that states that the scores measure interaction rather than the target, and that ``the other one'' refers to the instances not selected, raises RD to 73.5\% and overall SR to 57.4\%, 29.9 points above direct planning with the same model. The instance-anchored evidence is thus what lifts the VLM; reading it still requires stating the semantics that the symbolic programs encode. \method remains 6.8 points above this variant overall ($[1.3,12.7]$, $p=0.002$) and 6.9 points on RD ($[1.2,13.1]$); on NC the difference of 6.7 points is not significant ($[-4.1,17.9]$). The programs also decode a request in 0.1\,s instead of about 11\,s, return the same plan for the same evidence and provide the margin of Eq.~\eqref{eq:margin}. Attaching the frames to the plain evidence lowers NC SR to 29.7\%, because the model then overrides the events with its own reading of the images.

Table~\ref{tab:paired} tests the ensemble against every baseline at the level of activities, which are the units of the cross-validation. All differences are positive, and their 95\% intervals exclude zero except for the strict success of the explained-evidence variant, whose interval reaches zero. Against the seven VLM baselines that plan from frames or from the evidence with the plain prompt, \method is better on 46 to 59 of the 91 activities and worse on at most 11; against the explained-evidence variant, on 27 and worse on 8. Against the hand-crafted variant, which shares its registration, tracking and programs, most activities are tied, and 19 of the 26 activities that favour \method are NC activities.

\begin{table}[pos=!tbp]
  \centering\footnotesize
  \renewcommand{\arraystretch}{1.08}
  \setlength{\tabcolsep}{2pt}
  \caption{Paired comparison of \method (ensemble) with each baseline over the 91 implicit-intent activities.}
  \label{tab:paired}
  \begin{threeparttable}
    \begin{tabular*}{\tabw}{@{\extracolsep{\fill}}l c c c c@{}}
      \toprule
      & \multicolumn{3}{c}{Plan SR} & Strict \\
      \cmidrule(lr){2-4}\cmidrule(l){5-5}
      Baseline & $\Delta$ (95\% CI) & W/T/L\tnote{a} & $p$ & $\Delta$ (95\% CI)\tnote{b} \\
      \midrule
      8B, direct & $+41.1$\,$[31.2, 50.5]$ & 53/31/7 & $<\!10^{-9}$ & $+36.0$\,$[26.6, 45.3]$ \\
      32B, direct & $+36.7$\,$[27.0, 45.9]$ & 53/27/11 & $<\!10^{-6}$ & $+34.3$\,$[25.7, 43.1]$ \\
      8B, + overlays & $+33.4$\,$[24.2, 42.6]$ & 46/39/6 & $<\!10^{-7}$ & $+31.0$\,$[22.0, 40.2]$ \\
      32B, + overlays & $+35.4$\,$[26.4, 44.2]$ & 55/28/8 & $<\!10^{-9}$ & $+34.5$\,$[25.9, 42.9]$ \\
      InternVL3.5-8B & $+39.8$\,$[31.0, 48.1]$ & 59/27/5 & $<\!10^{-12}$ & $+38.0$\,$[29.0, 46.8]$ \\
      32B + evidence & $+33.6$\,$[23.7, 43.5]$ & 49/31/11 & $<\!10^{-6}$ & $+30.3$\,$[20.4, 40.7]$ \\
      \quad + frames & $+36.5$\,$[27.0, 45.9]$ & 51/30/10 & $<\!10^{-7}$ & $+33.4$\,$[24.0, 43.1]$ \\
      \quad explained & $+6.8$\,$[1.3, 12.7]$ & 27/56/8 & $0.002$ & $+5.3$\,$[0.0, 11.0]$ \\
      Hand-crafted & $+20.0$\,$[12.5, 27.9]$ & 26/61/4 & $<\!10^{-4}$ & $+8.6$\,$[3.1, 14.7]$ \\
      \bottomrule
    \end{tabular*}
    \begin{tablenotes}
      \item[a] \scriptsize Activities on which \method is better, tied or worse in plan SR.
      \item[b] \scriptsize Sign-test $p<10^{-4}$ for every VLM baseline except the explained-evidence variant ($p=0.014$; 24 activities better, 9 worse; the interval $[0.00, 10.99]$ reaches zero) and $p=0.052$ for the hand-crafted variant (16 activities better, 6 worse).
      \item \scriptsize $\Delta$: difference in percentage points over the 455 requests; intervals from 5{,}000 activity-cluster bootstrap resamples; $p$: two-sided sign test over activities with a non-zero difference, reported as an upper bound.
    \end{tablenotes}
  \end{threeparttable}
\end{table}

Fig.~\ref{fig:cases} shows examples. In Fig.~\ref{fig:cases}a, the frame logits rise at each object and destination cue, and the decoder recovers the two pairs in order; the 32B VLM moves all three objects to the tray. In Fig.~\ref{fig:cases}b, the two milk cartons are handled at different times and are identical at the end of the video; their evidence is kept apart because each mask is propagated from its own final-frame identity. Fig.~\ref{fig:cases}c shows a frequent NC failure: both objects are correct, but each is sent to the destination of the other. The object evidence of the two cream-cheese boxes overlaps in time, so the order of the object cues relative to the tray and basket cues is misread.

\begin{table}[pos=!tbp]
  \centering\footnotesize
  \renewcommand{\arraystretch}{1.08}
  \setlength{\tabcolsep}{2.6pt}
  \caption{Ablation and controls of \method (three seeds, \%).}
  \label{tab:ablation}
  \begin{threeparttable}
    \begin{tabular*}{\tabw}{@{\extracolsep{\fill}}l ccc cc cc@{}}
      \toprule
      & \multicolumn{2}{c}{NC} & RD & \multicolumn{2}{c}{All} & \multicolumn{2}{c}{$\Delta$ All\tnote{h}} \\
      \cmidrule(lr){2-3}\cmidrule(lr){4-4}\cmidrule(lr){5-6}\cmidrule(l){7-8}
      Variant & SR & Strict & Strict & SR & Strict & SR & Strict \\
      \midrule
      \method & $\mathbf{37.6}$ & 15.7 & 73.2 & 61.1 & 48.6 & -- & -- \\
      $-$ arm/head rays & 36.1 & 14.0 & $\mathbf{74.2}$ & 60.4 & 48.4 & $-0.7$ & $-0.1$ \\
      $+$ temporal conv.\tnote{a} & 33.5 & 14.7 & 72.7 & 59.2 & 47.8 & $-1.9$ & $-0.7$ \\
      $-$ program loss & 37.3 & 12.3 & 73.6 & 60.7 & 47.3 & $-0.4$ & $-1.2$ \\
      Event-level task loss\tnote{b} & 37.3 & $\mathbf{18.1}$ & 73.6 & $\mathbf{61.2}$ & $\mathbf{49.8}$ & $+0.1$ & $+1.2$\sig \\
      $-$ grammar decoding\tnote{c} & 21.2 & 8.9 & 73.2 & 54.1 & 45.6 & $-7.0$\sig & $-2.9$\sig \\
      $-$ backward tracking\tnote{d} & 34.2 & 13.3 & 70.1 & 57.2 & 45.8 & $-3.9$\sig & $-2.8$\sig \\
      Forward tracking\tnote{e} & 31.1 & 14.2 & 71.2 & 56.9 & 46.7 & $-4.2$\sig & $-1.8$ \\
      $-$ learning & 2.6 & 2.6 & 70.0 & 44.2 & 41.1 & $-16.9$\sig & $-7.5$\sig \\
      32 frames, evidence\tnote{f} & 35.2 & 11.6 & 71.7 & 57.9 & 45.9 & $-3.2$\sig & $-2.6$\sig \\
      32 frames, all stages\tnote{f} & 29.2 & 10.9 & 69.9 & 55.3 & 44.6 & $-5.8$\sig & $-4.0$\sig \\
      \midrule
      Recording-order blocks & 32.3 & 14.0 & 70.8 & 56.0 & 46.4 & $-5.1$\sig & $-2.1$ \\
      \midrule
      Relation model\tnote{g} & 19.7 & 7.9 & -- & -- & -- & -- & -- \\
      \bottomrule
    \end{tabular*}
    \begin{tablenotes}
      \item[a] \scriptsize Two temporal convolutions over the feature sequences before scoring.
      \item[b] \scriptsize Hamming loss between the events of a program and those of $\omega^{\star}$, used in both Eq.~\eqref{eq:augment} and Eq.~\eqref{eq:prog}.
      \item[c] \scriptsize Candidates with $s_c>0$, paired by the time of their peaks.
      \item[d] \scriptsize Final-frame boxes used for all frames. ``$-$ learning'' is the hand-crafted variant.
      \item[e] \scriptsize Detection on the first frame, forward propagation, identities from matching the tracks to the registered instances at the last frame.
      \item[f] \scriptsize Only the 32 frames given to the VLMs are used, in training and test: for the evidence, which is read at these frames from the full tracks, or for all stages, with tracking and features also computed on them; every other frame holds the nearest sampled frame.
      \item[g] \scriptsize Same features, folds and goal labels; a temporal CNN classifies every object--destination pair directly (NC only).
      \item[h] \scriptsize Paired difference to \method over the 455 requests, with outcomes averaged over seeds; \sig: 95\% activity-bootstrap interval excludes zero.
      \item \scriptsize Bold: best value in each column. Standard deviations over seeds are at most 2.7 points for NC SR and at most 3.4 points for the other columns.
    \end{tablenotes}
  \end{threeparttable}
\end{table}

\subsection{Ablation}

Table~\ref{tab:ablation} removes one component at a time. Grammar-constrained decoding is the most important element for NC: pairing thresholded candidates by the time of their peaks lowers NC SR from 37.6\% to 21.2\%, because spurious peaks and missed destinations break the object--destination structure. The paired differences in the last two columns, computed on seed-averaged outcomes per request, show which of these effects are larger than the variation across activities. Without grammar decoding overall SR drops by 7.0 points ($p=8\times10^{-5}$, sign test over activities). The program-level loss leaves SR unchanged and raises NC strict success from 12.3\% to 15.7\% (paired difference $3.4$ points, with a 95\% interval of $[0.00, 7.01]$ that reaches zero, $p=0.06$); it teaches the evidence network to suppress events that would add wrong operations, which the official metric does not penalize. An event-level Hamming loss to $\omega^{\star}$, used consistently in the search and in the hinge, gives a similar SR and a higher NC strict success (18.1\%), with a difference of $2.4$ points $[0.3, 4.8]$ that is not significant under the sign test ($p=0.29$); we keep the operation-level loss of Eq.~\eqref{eq:taskloss}, which was fixed before these runs and matches the evaluation. Backward tracking matters most for RD, where objects are picked up and moved during the selection: without it, RD strict success drops by 3.1 points, and overall SR by 3.9 points. The forearm and head rays and the temporal convolutions change no metric by more than the variation across activities.

\textbf{Controls.} Three further rows test alternatives to the design of \method. Backward anchoring could be replaced by the usual order of processing: detect the objects on the first frame, track them forwards, and assign identities by matching the tracks to the registered instances of the execution scene at the last frame. With the same detector, segmenter and counts, this matching recovers 94.4\% of the registered identities, and where both exist they agree with the backward tracks on the first frame for all but one of 1{,}220 instances. Its cost lies in the remaining instances: in 15\% of the NC requests and 5\% of the RD requests a goal object or container has no forward track, because it is missed on the first frame, lost on the way, or hidden in a container at the end. Overall SR drops by 4.2 points ($[-7.5,-1.2]$) and NC SR by 6.5 points. Anchoring at the execution scene guarantees a track for every instance that the plan can refer to. Second, the VLMs see 32 uniformly sampled frames, whereas \method reads evidence at 10\,Hz. We restricted \method to the same 32 frames in two ways. Reading the evidence only at these frames, from tracks computed on all frames, lowers overall SR by 3.2 points ($[-6.4,-0.3]$); running the whole pipeline on them, including mask propagation, keypoints and features, lowers it by 5.8 points ($[-9.2,-2.6]$), mostly on NC (8.4 points). The brief gestures thus profit from dense sampling, but with the same 32 frames \method still reaches 55.3\%, 24.5 points above the best VLM baseline. Third, the gain could stem from supervision on the task rather than from the program structure. A relation model with the same features, folds and goal labels, which classifies every object--destination pair with a temporal CNN, reaches 19.7\% SR on NC, 17.9 points below \method ($[-27.2,-9.4]$, $p=0.003$) and at the level of the VLMs. Training on outcomes alone is thus not sufficient; the frame evidence has to be read as whole programs, which the grammar decoder does exactly and which a 32B VLM can also do from the candidate events (Section~\ref{sec:results}). With recording-order blocked folds, overall SR decreases by 5.1 points and strict success by 2.1 points, so most of the gain transfers to activities recorded at other times.

\begin{table}[pos=!tbp]
  \centering\footnotesize
  \renewcommand{\arraystretch}{1.08}
  \setlength{\tabcolsep}{3pt}
  \caption{Episodic tasks without task-specific training: plan success rate (\%) on 800 requests.}
  \label{tab:episodic}
  \begin{threeparttable}
    \begin{tabular*}{\tabw}{@{\extracolsep{\fill}}l ccc c c@{}}
      \toprule
      Method & Imit. & Restore & Reversal & All [95\% CI] & Strict \\
      \midrule
      8B, direct & 37.6 & 0.4 & 5.8 & 15.4 {\scriptsize[10.5, 20.4]} & 14.1 \\
      32B, direct & 37.6 & 16.5 & 26.7 & 27.0 {\scriptsize[21.5, 32.6]} & 24.8 \\
      InternVL3.5-8B & 31.7 & 9.8 & 8.9 & 17.5 {\scriptsize[12.9, 22.5]} & 15.6 \\
      \method & $\mathbf{46.6}$ & $\mathbf{46.3}$ & $\mathbf{46.2}$ & $\mathbf{46.4}$ {\scriptsize[40.2, 52.2]} & $\mathbf{46.4}$ \\
      \bottomrule
    \end{tabular*}
    \begin{tablenotes}
      \item \scriptsize Imitation 290, restore 285 and reversal 225 requests; 8B and 32B are Qwen3-VL models. \method reuses registration and tracking with the episodic program of Section~\ref{sec:programs}. Paired difference of \method to the 32B model: $+19.4$ $[+12.2,+26.2]$; to the 8B model: $+31.0$ $[+23.6,+38.0]$; to InternVL3.5-8B: $+28.9$ $[+22.0,+35.4]$.
    \end{tablenotes}
  \end{threeparttable}
\end{table}

\begin{figure}[pos=!tbp]
  \centering
  \includegraphics[width=\textwidth]{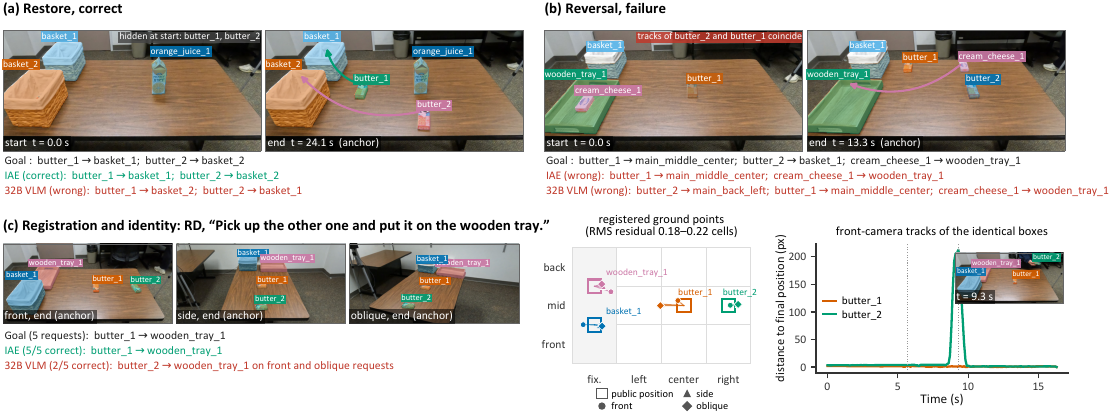}
  \caption{Episodic examples and instance registration. (a,b)~First and last frames of two episodic requests with the tracked instance masks and public identifiers that \method obtains at these frames; the anchor is the last frame, and arrows show the container destinations of the plan. In (a) both butter boxes are inside the baskets at the start, and each is assigned to the container nearest to its first visible position. In (b) the backward track of \texttt{butter\_2}, which starts inside \texttt{basket\_1}, coincides with the identical \texttt{butter\_1} in the first frame, so its place appears unchanged and its move is missed. (c)~One RD activity: last frames of the three cameras with the registered identifiers; registered ground points $\hat\pi(p(b))$ of all instances (filled markers, one shape per camera) against their public grid positions (open squares); distance of the two identical boxes from their final positions over time in the front camera, with the frame in which \texttt{butter\_2} is carried. Examples were drawn at random (fixed seed) from predefined groups: restoration or reversal requests from the front camera with at least two goal pairs, solved strictly by \method and not by the 32B VLM (48 requests); front-camera episodic requests that \method fails and whose goal involves a container (124); RD activities with two instances of one category, successful registration on all three cameras and a moved instance (29). Frames are cropped at the same height for all cameras.}
  \label{fig:qual2}
\end{figure}

\subsection{Episodic tasks}

Table~\ref{tab:episodic} applies the same registration and tracking to restoration, reversal and imitation, with the episodic program and no training. Both VLMs imitate equally well (37.6\%), since the final positions are visible at the end of the video. Restoration and reversal require remembering where each instance started: the 8B model almost never solves them (0.4\% and 5.8\%), and the 32B model reaches 16.5\% and 26.7\%; InternVL3.5-8B reaches 9.8\% and 8.9\%, and 31.7\% on imitation. \method solves the three tasks at the same rate of about 46\%. Overall it is 19.4 points above the 32B model (95\% interval $[+12.2,+26.2]$, better on 79 activities and worse on 28, $p=8\times10^{-7}$) and 31.0 points above the 8B model ($[+23.6,+38.0]$, $p=10^{-14}$). Its success equals its strict success: every successful plan, which moves only instances whose place changed, also left all other objects in place. The remaining failures are dominated by objects taken out of or put into containers, for which the instance is hidden at one end of the video, and by confusions between two adjacent baskets.

Fig.~\ref{fig:qual2} shows what the programs operate on. In Fig.~\ref{fig:qual2}a both butter boxes are hidden in the baskets at the start of the video; the program assigns each to the container nearest to its first visible position and restores both, whereas the 32B VLM swaps the two identical boxes. Fig.~\ref{fig:qual2}b shows a failure mode behind container errors: \texttt{butter\_2} starts inside \texttt{basket\_1}, its backward track collapses onto the identical \texttt{butter\_1}, and the program sees no change of place. Fig.~\ref{fig:qual2}c shows that registration assigns the same identifiers to the two identical boxes in all three cameras, with root-mean-square residuals of 0.18 to 0.22 grid cells, and that the track of \texttt{butter\_2} records it being lifted and put back about 9\,s into the video. On the five requests of this activity, \method moves the other box, \texttt{butter\_1}, every time; the 32B VLM does so on the two side-camera requests and moves the handled box on the front and oblique ones.

\begin{figure}[pos=!tbp]
  \centering
  \includegraphics[width=\textwidth]{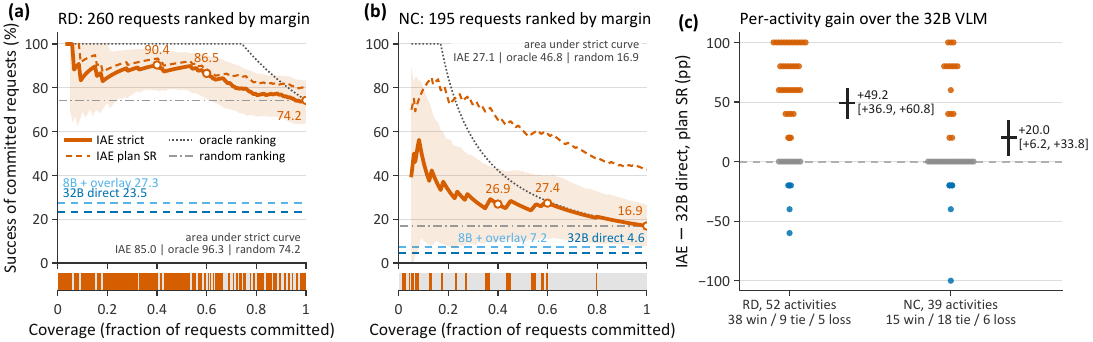}
  \caption{Reliability and consistency of \method on the implicit-intent tasks. (a,b)~Success of the committed requests when requests are ranked by the confidence of the three-seed ensemble (NC: program margin $\mu$; RD: selection margin $\min_c|s_c|$); shaded bands are 95\% activity-bootstrap intervals of strict success, and curves start at ten committed requests. The strip under each panel marks every request in ranked order, filled when strictly successful. Dotted and dash-dotted grey curves are the oracle ranking (all strictly successful requests first) and a random ranking; the areas under the three strict curves are given in each panel. Dashed coloured lines give the strict success of two direct baselines, which provide no confidence. (c)~Per-activity difference in plan SR between \method and Qwen3-VL-32B direct planning; black bars give the mean and its 95\% bootstrap interval.}
  \label{fig:selective}
\end{figure}

\begin{table}[pos=!tbp]
  \centering\footnotesize
  \renewcommand{\arraystretch}{1.08}
  \setlength{\tabcolsep}{2.5pt}
  \caption{Plan success rate (\%) by camera and spatial reference.}
  \label{tab:views}
  \begin{threeparttable}
    \begin{tabular*}{\tabw}{@{\extracolsep{\fill}}l ccc cc@{}}
      \toprule
      & \multicolumn{3}{c}{Camera} & \multicolumn{2}{c}{Reference} \\
      \cmidrule(lr){2-4}\cmidrule(l){5-6}
      Method & Front & Side & Obl. & Camera & Person \\
      \midrule
      \multicolumn{6}{@{}l}{\textit{Nonverbal cue}} \\
      32B, direct & 28.2 & 14.1 & 28.2 & 25.6 & 20.5 \\
      8B, direct + overlays & 26.9 & 23.1 & 28.2 & 25.6 & 25.6 \\
      \method (ensemble) & $\mathbf{46.2}$ & $\mathbf{41.0}$ & $\mathbf{38.5}$ & $\mathbf{43.6}$ & $\mathbf{41.9}$ \\
      \midrule
      \multicolumn{6}{@{}l}{\textit{Reference disambiguation}} \\
      32B, direct & 34.6 & 28.8 & 28.8 & 31.7 & 30.8 \\
      8B, direct + overlays & 46.2 & 25.0 & 30.8 & 36.5 & 33.3 \\
      \method (ensemble) & $\mathbf{86.5}$ & $\mathbf{75.0}$ & $\mathbf{78.8}$ & $\mathbf{80.8}$ & $\mathbf{80.1}$ \\
      \bottomrule
    \end{tabular*}
    \begin{tablenotes}
      \item \scriptsize NC: 78, 78 and 39 requests from the front, side and oblique cameras; 78 camera-centred and 117 person-centred requests. RD: 104, 104 and 52; 104 and 156.
    \end{tablenotes}
  \end{threeparttable}
\end{table}

\subsection{Cameras, verification and selective prediction}

\textbf{Cameras and spatial reference.} Table~\ref{tab:views} breaks down the implicit-intent results by camera and by the spatial frame in which the request is phrased. \method improves on every camera. Success is highest on the front camera for both tasks, and the oblique camera is the hardest for NC. \method is nearly insensitive to the spatial frame of the request (differences of at most 1.7 points), whereas the 32B VLM loses 5.1 points on person-centred NC requests.

\textbf{Can a VLM verify ambiguous destinations?} Among goal pairs, destination errors are the largest error type on NC (Section~\ref{sec:errors}). Among the 327 objects that the ensemble selects correctly, 214 receive the correct destination; 55 of the errors confuse the tray with the basket, 25 choose the wrong drawer and 24 choose a container instead of a table region. We tested whether a VLM can resolve such cases, using the programs decoded by the seed-0 model, which assign 203 correct destinations. For every decoded destination event whose competing destinations lie within 1.5 logits (124 events), we showed 7 frames around the event with the competitors marked by letters and asked the VLM which one the person indicated. The 8B and 32B models changed 51 and 55 of these destinations, but the number of correct destinations rose only to 205 and 204, and NC SR decreased by 4.6 and 4.1 points. The failures arise from a real ambiguity in the image: a downward point at the tray from behind the table places the fingertip over the basket, and the two VLMs we tested follow the fingertip position.

\textbf{Selective prediction.} The decoding margin and the selection margin rank requests by reliability (Fig.~\ref{fig:selective}a,b). On RD, committing to the 60\% most confident requests raises strict success of the committed requests from 74.2\% to 86.5\%, and to 90.4\% at 40\% coverage. On NC the margin separates a small confident group, of which about half is strictly correct (53.3\% of the 15 most confident requests), from the rest: strict success is 33.3\% at 20\% coverage and 27.4\% at 60\%, because confidently decoded programs with a wrong destination or an extra object remain. The areas under the strict curves summarize this difference: on RD the margin recovers 49\% of the gap between a random ranking (74.2) and the oracle (96.3), reaching 85.0, whereas on NC it recovers 34\% (27.1 between 16.9 and 46.8). The NC margin is therefore a ranking signal rather than a guarantee for execution. Because \method outperforms the VLMs on both tasks, falling back to the 32B VLM for the least confident 10--50\% of requests lowers overall SR (from 64.2\% to between 62.4\% and 51.2\%).

\textbf{Consistency across activities.} Fig.~\ref{fig:selective}c shows that the gain over the 32B VLM is not driven by a few activities. On RD, \method is better on 38 of 52 activities and worse on 5; on NC, it is better on 15 of 39 and worse on 6, while 18 activities are tied, 16 of which neither method solves for any request. These unsolved activities mark the part of NC that neither the VLM nor the current evidence can read.

\begin{figure}[pos=!tbp]
  \centering
  \includegraphics[width=\textwidth]{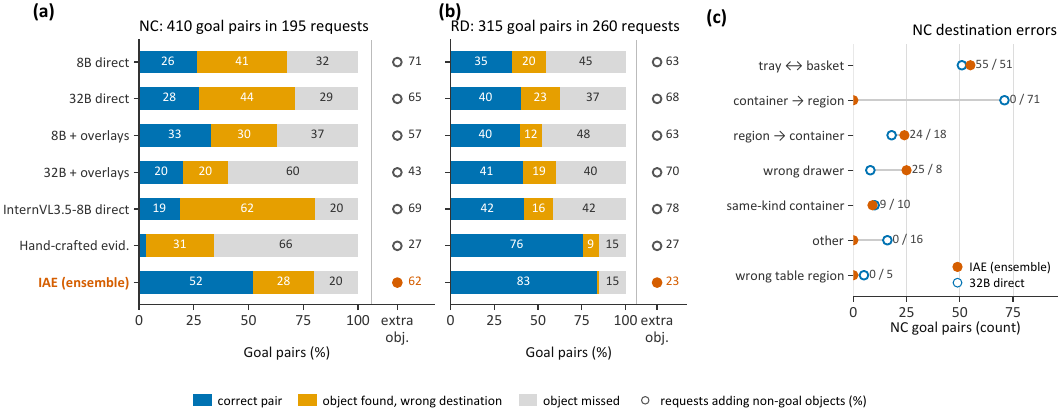}
  \caption{Where plans fail on the implicit-intent tasks. (a,b)~Outcome of every goal pair: the object is moved to its goal destination (correct pair), moved elsewhere (wrong destination) or not moved (missed). Circles give the share of requests whose plan additionally moves an object that is not in the goal, which the official metric does not penalize but strict success does. (c)~Types of wrong destinations on NC for \method (ensemble, filled) and Qwen3-VL-32B direct planning (open); numbers are counts for \method / 32B.}
  \label{fig:errors}
\end{figure}

\subsection{Error analysis}
\label{sec:errors}

Fig.~\ref{fig:errors} decomposes the plans into the outcomes of individual goal pairs. On RD, \method moves 83\% of the goal objects to the right container and almost never chooses a wrong destination (1.6\%), because the destination is fixed by the instruction once the object is selected; the remaining errors are selection errors. The VLMs miss 37--48\% of the RD goal objects and add a non-goal object in 63--78\% of the requests.

On NC, \method places 52\% of the goal pairs correctly, against 19--33\% for the VLMs. It misses 20\% of the goal objects, as many as InternVL3.5-8B (20\%) and fewer than the Qwen3-VL baselines (29--60\%); InternVL3.5-8B instead sends 62\% of the goal objects to a wrong destination. Two error sources remain. First, 28\% of the goal pairs receive a wrong destination; Fig.~\ref{fig:errors}c shows that these are dominated by tray--basket confusions (55 pairs, against 51 for the 32B VLM) and by wrong drawers, which are both cases of targets that are adjacent in the image and separated mainly in depth. The 32B VLM instead most often places an object on a table region when the goal is a container (71 pairs), an error that \method does not make. Second, 62\% of the NC plans of \method move at least one non-goal object. Averaged over requests, 46.9\% of its NC operations are goal operations and it produces 59.3\% of the goal operations, with 0.78 non-goal objects per plan, against 21.2\%, 32.0\% and 0.90 for the 32B VLM. These extra objects come from spurious object events that the decoder accepts because they raise the score of a valid program, which $\lambda=0$ does not penalize, and they account for the gap between plan SR (42.6\%) and strict success (16.9\%), since a request that meets its goal fails strict success only when a non-goal object is moved. Section~\ref{sec:sensitivity} shows that the decoding cost trades the two metrics against each other.

\textbf{Which error dominates?} To locate the bottleneck, we decoded the ensemble logits once more with information that \method never receives, for analysis only. Given the set of goal objects, with destinations and pairing still read from the evidence, NC strict success rises from 16.9\% to 48.7\%, so most strict failures are extra or missing objects. Even then, 39\% of the goal objects (95 of 243 over the NC videos) are sent to a wrong destination, whereas limiting the destination events to the goal destinations raises NC SR only from 42.6\% to 43.6\%. The remaining destination errors are therefore mostly confusions among the destinations of the same goal, as in Fig.~\ref{fig:cases}c, rather than choices of a destination outside the goal.

\begin{figure}[pos=!tbp]
  \centering
  \includegraphics[width=\textwidth]{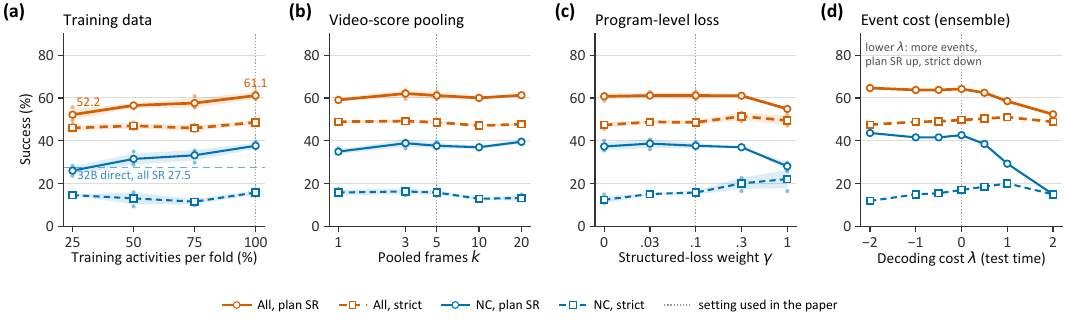}
  \caption{Data efficiency and sensitivity of \method on the implicit-intent tasks (plan SR, solid; strict success, dashed; all 455 requests in orange, NC in blue). (a--c)~Retraining with a fraction of the training activities of each fold, a different number $k$ of frames pooled into the video score, and a different weight $\gamma$ of the program-level loss; lines are means over three seeds, bands one standard deviation, and small dots the individual seeds. (d)~Test-time decoding cost $\lambda$ applied to the frame logits of the three-seed ensemble without retraining; RD does not depend on $\lambda$. Dotted lines mark the settings used elsewhere in the paper.}
  \label{fig:sensitivity}
\end{figure}

\subsection{Data efficiency and sensitivity}
\label{sec:sensitivity}

\textbf{Training data.} Fig.~\ref{fig:sensitivity}a retrains the evidence network on 25\%, 50\% and 75\% of the training activities of each fold, drawn separately for NC and RD. With a quarter of the activities, about eight NC and ten RD activities per fold, overall SR is 52.2\%, about twice that of the 32B VLM, and RD alone reaches 71.8\%. NC depends more on the amount of supervision: its SR rises from 26.0\% with a quarter of the activities to 31.5\%, 33.2\% and 37.6\%, which suggests that more pointing demonstrations would further improve the evidence.

\textbf{Hyperparameters.} The number of pooled frames hardly matters for plan SR (59.0--62.1\% for $k$ from 1 to 20), while pooling more than five frames lowers NC strict success from 15.7--16.2\% to 12.8--13.2\%, possibly because a long window favours candidates that stay near the hand over candidates that are briefly indicated (Fig.~\ref{fig:sensitivity}b). The weight of the program-level loss trades NC plan SR against strict success (Fig.~\ref{fig:sensitivity}c): weights up to $\gamma=0.3$ raise NC strict success from 12.3\% to 20.0\% at nearly unchanged SR (36.9--38.6\%), whereas $\gamma=1$ suppresses so many events that NC SR drops to 28.0\%. We keep the weight $\gamma=0.1$ that was fixed before the experiments rather than selecting it on the test folds.

\textbf{Decoding cost.} The event cost $\lambda$ of Eq.~\eqref{eq:decode} controls how many events the decoder accepts (Fig.~\ref{fig:sensitivity}d). Lowering it to $\lambda=-2$ raises NC plan SR to 43.6\% but lowers NC strict success to 11.8\%, because the additional events mostly add objects to the plan, which the official metric ignores. Raising it to $\lambda=1$ gives the highest strict success (20.0\% on NC, 51.0\% overall) at a plan SR of 58.5\%. We also selected $\lambda$ by the number of videos whose decoded operations equal the goal. Decoding the training videos themselves picks $\lambda=1$ in all 15 folds of the three seeds; a nested cross-validation, which retrains the network on three quarters of each training fold and decodes the remaining quarter, picks $0.5$ or $1$ in 13 of the 15 folds. On the untouched test folds, the nested choice changes NC strict success by $+0.9$ points ($[-2.7,5.1]$) and lowers NC SR by 6.2 points ($[-9.9,-2.9]$). We keep $\lambda=0$, which charges nothing for an event, and report both metrics throughout.

\begin{table}[pos=!tbp]
  \centering\footnotesize
  \renewcommand{\arraystretch}{1.08}
  \setlength{\tabcolsep}{2.5pt}
  \caption{Dependence on the public layout used for registration (\%, three seeds).}
  \label{tab:layout}
  \begin{threeparttable}
    \begin{tabular*}{\tabw}{@{\extracolsep{\fill}}l c cc cc c@{}}
      \toprule
      & & NC & RD & \multicolumn{2}{c}{All} & Episodic \\
      \cmidrule(lr){3-3}\cmidrule(lr){4-4}\cmidrule(lr){5-6}\cmidrule(l){7-7}
      Layout at test time & IDs changed\tnote{a} & SR & SR & SR & Strict & SR \\
      \midrule
      Exact layout & 0.0 & $\mathbf{37.6}$ & $\mathbf{78.7}$ & $\mathbf{61.1}$ & $\mathbf{48.6}$ & $\mathbf{46.4}$ \\
      Noise $\sigma=0.25$ cell & 4.5 & 37.1 & 78.5 & 60.7 & 48.2 & 44.0 \\
      Noise $\sigma=0.5$ cell & 15.3 & 35.2 & 72.2 & 56.3 & 44.5 & 38.0 \\
      Noise $\sigma=1$ cell & 34.2 & 34.0 & 60.1 & 48.9 & 37.1 & 30.7 \\
      Random identities\tnote{b} & 51.4 & 32.8 & 49.2 & 42.2 & 30.5 & 27.1 \\
      \midrule
      32B, direct & -- & 22.6 & 31.2 & 27.5 & 15.4 & 27.0 \\
      \bottomrule
    \end{tabular*}
    \begin{tablenotes}
      \item[a] \scriptsize Videos whose identifier assignment differs from the exact-layout registration.
      \item[b] \scriptsize Registration map kept, identifiers permuted at random among detections of the same category.
      \item \scriptsize Networks are trained on exact-layout features and applied to perturbed test inputs; means over three seeds (the noise draw follows the seed). Standard deviations are at most 4.0 points.
    \end{tablenotes}
  \end{threeparttable}
\end{table}

\textbf{Dependence on the public layout.} Registration uses the public layout, which the baselines receive as well, but a deployed system would know it only approximately. Table~\ref{tab:layout} perturbs the layout at test time: Gaussian noise is added to the canonical coordinates of every object before registration, which changes the fitted map and, for some videos, the assignment of identifiers; in the last condition identifiers are permuted at random within each category while the map is kept. The networks are not retrained. With noise of a quarter cell, 4.5\% of the videos change their assignment and overall SR drops by 0.4 points. Larger noise degrades mostly the tasks that depend on identity: RD falls from 78.7\% to 60.1\% at one cell and the episodic tasks from 46.4\% to 30.7\%, whereas NC loses 3.6 points. With random identities, the episodic programs reach 27.1\%, the level of the 32B VLM, which confirms that their success rests on identity bookkeeping; on the implicit-intent tasks overall SR is 42.2\%, still above every VLM baseline in Table~\ref{tab:main}.

\textbf{Cost.} Table~\ref{tab:cost} lists the computation for two videos, about 2.1\,s per second of video. Over 20 further NC and RD videos drawn at random (12.8 to 38.0\,s long), the same stages take 44.8\,s per video in the median (90th percentile 56.3\,s), or 2.06\,s per second of video (90th percentile 2.17\,s); this measurement ran on a machine shared with other jobs. Per-frame person detection and mask propagation dominate; perception is computed once per video, and all requests on that video share it. The evidence network and the programs take less than half a second. Direct planning costs 6\,s per request with the 8B model and 18\,s with the 32B model on average, and is repeated for every request.

\begin{table}[pos=!tbp]
  \centering\footnotesize
  \renewcommand{\arraystretch}{1.08}
  \setlength{\tabcolsep}{3pt}
  \caption{Computation for two test videos of 34.5\,s and 33.3\,s (one RTX 5880 Ada GPU; keypoints on CPU).}
  \label{tab:cost}
  \begin{threeparttable}
    \begin{tabular*}{\tabw}{@{\extracolsep{\fill}}l l c c@{}}
      \toprule
      Stage & Model & \multicolumn{2}{c}{Time (s)} \\
      \midrule
      Anchor-frame detection & Grounding DINO-T & 1.1 & 0.3 \\
      Layout registration & least squares & $<$0.01 & $<$0.01 \\
      Mask propagation & SAM~2.1-T & 20.2 & 19.0 \\
      Per-frame person detection & Grounding DINO-T & 36.1 & 35.3 \\
      Hand keypoints & MediaPipe Hands & 8.0 & 10.2 \\
      Body keypoints & RTMPose-m & 4.4 & 4.5 \\
      Features, network, programs & -- & 0.1 & 0.1 \\
      \midrule
      \method, total per video & & 70.0 & 69.4 \\
      \bottomrule
    \end{tabular*}
    \begin{tablenotes}
      \item \scriptsize Stages run sequentially on an otherwise idle machine; perception is shared by all requests on a video (on average 1.7). Direct planning takes 6.2\,s (8B) and 18.0\,s (32B, FP8) per request on average over the 455 requests.
    \end{tablenotes}
  \end{threeparttable}
\end{table}

\section{Discussion and limitations}
\label{sec:discussion}

The results separate two parts of behavior-grounded planning that a VLM performs jointly. Identity bookkeeping, knowing which instance was touched or moved and where it came from, is solved to a large extent by registration and tracking alone: the hand-crafted variant reaches 75.4\% on RD, and the episodic programs reach 46.4\% without learning, whereas the VLMs lose track of identical instances. Reading pointing gestures is the harder part. It benefits from learned evidence and from decoding whole programs, and it remains the main source of errors, with 16.9\% strict success on NC. Once the evidence is anchored to instances, a large VLM prompted with its semantics plans from it at 57.4\% SR, against 27.5\% from the frames; the explicit programs add 6.8 points, run about a hundred times faster and are deterministic. The controls indicate that neither denser sampling nor outcome supervision alone explains the gain: \method restricted to the frames of the VLMs at every stage stays far above them, and a relation model trained on the same labels without program decoding does not. They also show where backward anchoring helps: forward tracking assigns the same identities when it keeps an instance, but loses instances that are not visible throughout the video.

Five limitations bound these conclusions. First, registration relies on the public layout of the execution scene. Small errors in it are tolerated, but with errors of one grid cell a third of the videos change their identity assignment, and overall SR drops by 12 points (Table~\ref{tab:layout}); when two identical instances share a region, or the layout is unavailable, a different anchor would be required. Second, the evidence is two-dimensional. Destinations that are aligned in depth, such as a tray in front of a basket, cannot always be separated from a single view; the two VLMs we asked to verify these destinations did not improve them. Metric depth or a second view could. Third, the plans are evaluated by symbolic execution in the benchmark's simulator. The results show which objects a robot should move and where, not whether it can grasp and place them; executing the plans of \method, of the strongest baseline and of the goal with one fixed controller would separate planning from execution failures, and is left to future work. Fourth, all experiments use one benchmark and one set of recording rooms. The recording-order split reduces but does not remove the overlap between training and test conditions, and a test on new rooms, participants and layouts with all parameters frozen is needed to establish transfer. Fifth, on NC more than 60\% of the plans still move an extra object. The decoding cost and the weight of the program loss trade this against plan success, and a cost selected by nested cross-validation does not raise strict success on new activities while it lowers plan success; reducing spurious object events without losing true ones needs better evidence rather than a different threshold.

\section{Conclusion}
\label{sec:conclusion}

We presented instance-anchored interaction evidence, which binds human interactions in a demonstration video to the object instances that a robot will manipulate. Registering the execution scene to its public layout and tracking every instance backwards makes identity explicit; hand and forearm geometry relative to these instances forms the evidence; and a network trained only from task outcomes, together with grammar-constrained program decoding, turns pointing gestures into object--destination plans. On WatchAct, with plans scored by symbolic execution, this raises plan success on implicit-intent tasks from 27.5\% for a 32B VLM to 64.2\% (strict success from 15.4\% to 49.7\%), and raises success on restoration, reversal and imitation from 27.0\% to 46.4\% without task-specific training. Suppressing spurious object events, which cause most strict failures on pointing tasks, and resolving destinations that are ambiguous in depth are the main remaining steps.

\section*{CRediT authorship contribution statement}
\textbf{Xinliang Xiao:} Conceptualization, Methodology, Software, Investigation, Formal analysis, Visualization, Writing -- original draft. \textbf{Bowen Yang:} Methodology, Software, Validation, Data curation, Investigation, Writing -- review \& editing. \textbf{Wenjing Zhang:} Validation, Visualization, Writing -- review \& editing. \textbf{Li Yang:} Conceptualization, Validation, Supervision, Funding acquisition, Writing -- review \& editing. \textbf{Wei Zhou:} Conceptualization, Methodology, Supervision, Project administration, Funding acquisition, Writing -- review \& editing.

\section*{Declaration of competing interest}
The authors declare that they have no known competing financial interests or personal relationships that could have appeared to influence the work reported in this paper.

\section*{Data availability}
The WatchAct benchmark used in this study is publicly available~\cite{watchact}. The code, the assignment of activities to cross-validation folds and the per-request predictions of all methods are available at \url{https://github.com/WeiZhou96/iae-watchact}.

\section*{Declaration of generative AI and AI-assisted technologies in the writing process}
During the preparation of this work the authors used Claude (Anthropic) in order to improve the language of the manuscript. After using this tool, the authors reviewed and edited the content as needed and take full responsibility for the content of the published article.

\section*{Funding}
This work was supported by the Jiangsu Provincial Major Science and Technology Program (No.~BG2025043), the National Natural Science Foundation of China (No.~52502402), the Basic Research Program of Jiangsu Province (No.~BK20251462), and the Natural Science Foundation of Jiangsu Province (No.~BK20232038).

\appendix

\section{Prompts of the VLM baselines}
\label{app:prompts}

All VLM baselines decode greedily. Each frame is preceded by the line \texttt{FRAME $i$ TIME $t$ seconds}; the instruction is given without the benchmark prefix that asks to start from the final scene, and \texttt{\{scene\}} is a JSON object with the public object list, their initial states, the allowed regions and the region descriptions, all in the spatial frame of the request. Predicted regions are mapped back to the canonical frame before scoring.

\textbf{Direct planning (system and user prompt).}
\begin{lstlisting}
SYSTEM: You interpret a human demonstration video for a robot that will act AFTER the video, starting from the final scene. The person may select objects or locations by pointing, touching, picking up, or moving them. Use only the supplied frames and public scene information. Return one JSON object following the user's schema.
USER: Task instruction: {instruction}
Infer every object placement the robot should perform. Return one JSON object {"pairs":[{"object_id":"exact public object ID","destination_region":"exact public region ID"}]} with one to three pairs. Always return your best-supported answer; an empty answer counts as a failure. Region IDs follow the spatial reference described below.
PUBLIC_SCENE
{scene}
\end{lstlisting}
For the episodic tasks, ``with one to three pairs'' is replaced by ``with every object that must be moved (one to eight pairs)''. The overlay variants prepend: \emph{Frames are annotated: coloured boxes mark the tracked objects and are labelled with their exact public object IDs; yellow lines show the detected hand skeletons and the forearm direction.}

\textbf{Evidence input.} The evidence variants prepend the following block to the user prompt of direct planning.
\begin{lstlisting}
STRUCTURED_EVIDENCE (computed from the video by a perception system that registers every object of the final scene to its public ID, tracks it through the video and scores hand-object interactions; higher scores mean stronger evidence that the person indicated or handled that candidate):
{evidence}
\end{lstlisting}
For NC, \texttt{\{evidence\}} lists the candidate events of Eq.~\eqref{eq:events} in temporal order, one per line, as \texttt{t=6.5s destination wooden\_cabinet\_1\_top\_region score 3.06} or \texttt{t=7.6s object cream\_cheese\_1 score 1.32}. For RD, it lists the video-level score $s_c$ of every movable instance and of every container, and the distance in grid cells from the person's median table position to every container.

\textbf{Evidence input with explained semantics.} This variant drops the clause on higher scores from the parenthesis above and appends, after the evidence, ``How to read this evidence:'' followed by, for NC,
\begin{lstlisting}
each event is a moment at which the person indicated a candidate with a hand; a higher score means stronger evidence that the candidate was indicated at that time, and events with low scores are likely spurious. A demonstration usually indicates one or more objects and then the place where they should go, possibly several times.
\end{lstlisting}
and, for RD,
\begin{lstlisting}
an interaction score is evidence that the person touched, pointed at or handled that instance during the video; it is NOT the probability that the instance must be moved. Decide the target objects from the instruction: if it refers to the selected/indicated/handled object(s), the targets are the instances with high scores; if it refers to "the other one" or "the others", the targets are the instances of the same kind that the person did NOT select, i.e. the ones with low scores. A container score is evidence that the person indicated that container.
\end{lstlisting}

\textbf{Destination verification.}
\begin{lstlisting}
SYSTEM: You judge where a person is pointing in short video clips. Answer with a single letter.
USER: The frames show a person indicating a destination with a pointing gesture. Candidate destinations are marked with letters {letters}. Which marked destination is the person pointing at? Consider the direction of the finger and arm, not only which mark is closest to the fingertip in the image. Answer with exactly one letter.
\end{lstlisting}

\section{Relation model}
\label{app:relation}

The same-supervision relation model of Table~\ref{tab:ablation} embeds every hand--candidate feature vector $\phi_{t,h,c}$ with a shared two-layer perceptron (48 units, ReLU) and takes the maximum over the visible hands, with zeros for frames without a hand. For every pair of a movable object and a destination candidate, the two embedding sequences are concatenated (96 channels) and passed through two one-dimensional convolutions over time (64 channels, kernel 9, dilations 1 and 2, ReLU), followed by maximum and mean pooling over time and a linear output. It is trained on the NC videos of the training folds with a class-balanced binary cross-entropy over all pairs, a pair being positive when it is a goal operation, using Adam (learning rate $3\times10^{-3}$, weight decay $10^{-4}$) for 150 full-batch epochs, as the evidence network. At test time every object takes its best destination, and objects whose best pair logit exceeds a threshold $\theta$ are kept (at most three, at least one); $\theta\in\{-3,-2,-1,-0.5,0,0.5,1,2,3\}$ is chosen on the training videos of each fold by the number of videos whose operations equal the goal.

\bibliographystyle{elsarticle-num}
\bibliography{iae_refs}

\end{document}